\documentclass[sigconf]{acmart}
\usepackage{subcaption}
\usepackage{multirow}
\AtBeginDocument{%
  }

\setcopyright{acmlicensed}
\copyrightyear{2026}
\acmYear{2026}
\acmDOI{XXXXXXX.XXXXXXX}
\acmConference[Conference acronym 'XX]{Make sure to enter the correct
  conference title from your rights confirmation email}{June 03--05,
  2018}{Woodstock, NY}
\acmISBN{978-1-4503-XXXX-X/2018/06}

\begin{document}

\title{ADMC: Attention-based Diffusion model for Missing modalities Completion}

\author{Yuhan Li}
\email{liyuhan@nudt.edu.cn}
\affiliation{
  \institution{National University of Defense Technology}
  \city{ChangSha}
  \country{China}}

\author{Wei Zhang}
\authornote{Both authors contributed equally to this research.}
\email{zhangwei23@nudt.edu.cn}
\orcid{0009-0009-8079-3349}
\affiliation{
  \institution{The Fifth Electronic Research Institute of MIIT}
  \city{GuangDong}
  \country{China}}

\author{Juan Chen}
\email{chenjuan@ict.ac.cn}
\affiliation{
  \institution{University of Chinese Academy of Sciences}
  \city{BeiJing}
  \country{China}}

\author{Jiangjia Yan}
\email{m202510806@xs.ustb.edu.cn}
\affiliation{
  \institution{University of Science and Technology Beijing}
  \city{Bei Jing}
  \country{China}}

\author{Peng Xiangli}
\email{xianglipeng@163.com}
\orcid{0009-0007-4182-5630}
\affiliation{
  \institution{Hunan University}
  \city{ChangSha}
  \country{China}}

\author{Liangze Yin}
\authornote{Corresponding author.}
\email{yinliangze@163.com}
\affiliation{
  \institution{National University of Defense Technology}
  \city{Changsha}
  \country{China}
}


\begin{abstract}

Multimodal emotion and intent recognition is essential for automated human-computer interaction, and it aims to analyze users' speech, text, and visual information to predict their emotions or intent. One of the significant challenges is that missing modalities due to sensor malfunctions or incomplete data. Traditional methods that attempt to reconstruct missing information often suffer from over-coupling and imprecise generation processes, leading to suboptimal outcomes. To address these issues, we introduce an Attention-based Diffusion model for Missing Modalities feature Completion (ADMC). Our framework independently trains feature extraction networks for each modality, preserving their unique characteristics and avoiding over-coupling. The Attention-based Diffusion Network (ADN) generates missing modality features that closely align with authentic multimodal distribution, enhancing performance across all missing-modality scenarios. Moreover, ADN's cross-modal generation offers improved recognition even in full-modality contexts. Our approach achieves state-of-the-art results on the IEMOCAP and MIntRec datasets, with up to 9.4\% improvement in Average Weighted Accuracy, demonstrating its effectiveness.

\end{abstract}



\maketitle

\section{Introduction}

\begin{figure}[!t]
\centering
\includegraphics[width=1\linewidth]{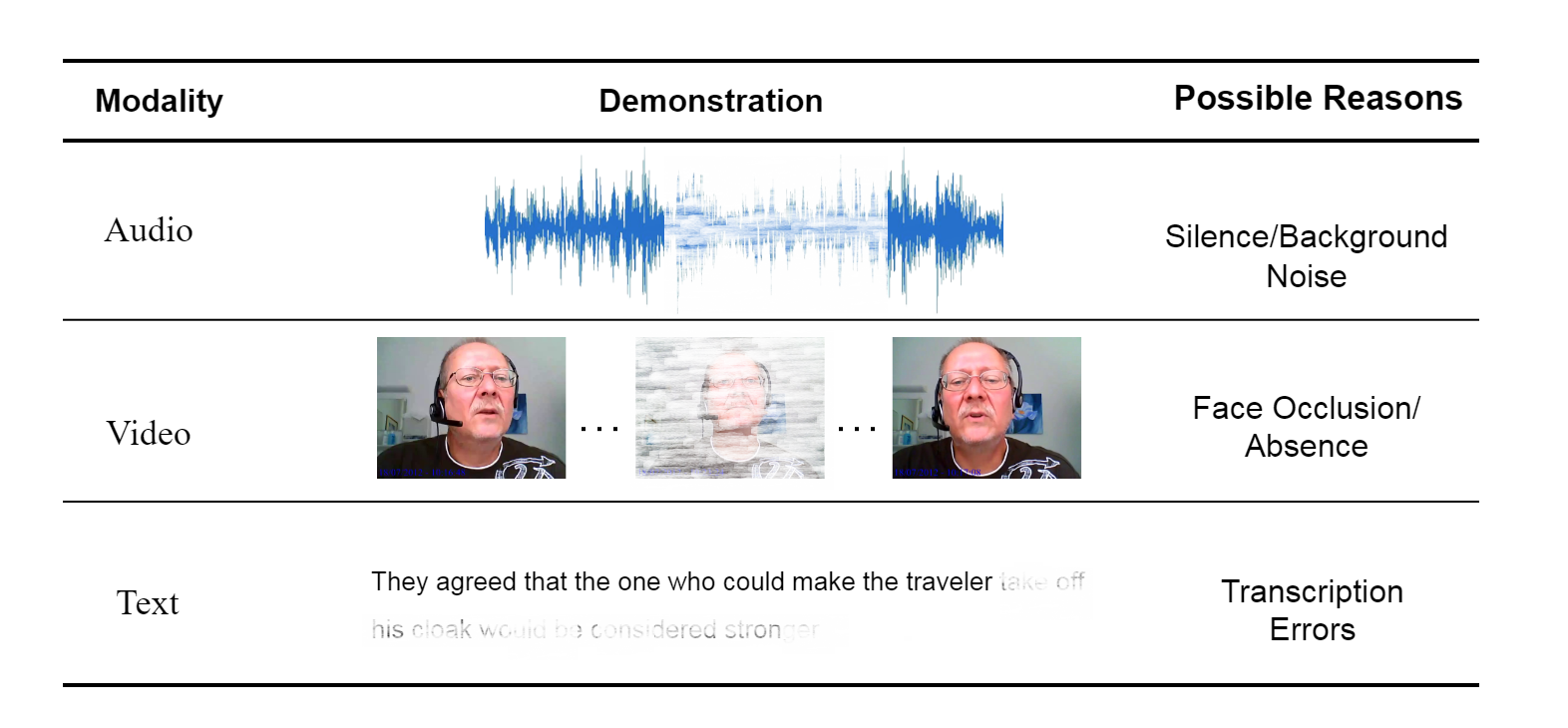}
\caption{In real-world human-computer interaction, modality missingness frequently occurs due to factors such as speech recognition errors, user silence, or visual occlusion. These issues significantly impact the performance of multimodal recognition and pose challenges to the effective fusion of multimodal information.
}
\label{fig_1}
\end{figure}

Multimodal emotion and intent recognition \cite{EmoCaps,ref5,ref6} is pivotal for automated human-computer interaction, as it seeks to interpret and identify human emotions and intentions through various modalities such as speech content, voice tones, and facial expressions. However, real-world applications often face the challenge of missing modalities due to incomplete data collection or sensor malfunctions, including speech recognition errors, user silence, or visual occlusion, as shown in Fig.~\ref{fig_1}. These issues significantly complicate multimodal recognition.

To tackle missing modalities, several existing studies \cite{adversarial, CRA, MCTN, MMIN, TATE, EMMR} focus on leveraging available modalities to reconstruct the missing information, thus mitigating the adverse effects on model performance caused by modality loss. Approaches like those in \cite{MCTN, TransM} utilize modality translation to create a unified representation, but this requires separate reconstruction models for each missing modality scenario, leading to increased costs and inconvenience. To address this, \cite{MMIN} proposed a unified framework capable of handling all missing modalities within a single model. This framework, known as the Missing Modality Imagination Network (MMIN), employs cascade residual auto-encoders \cite{CRA} and cycle consistency learning to generate the missing modalities. Building upon MMIN, \cite{IF-MMIN} enhances model performance by introducing modality-invariant features to bridge modality gaps.

Despite advancements, significant challenges persist. Current methods often attempt to recover the multimodal joint representation by jointly training feature extraction and classification networks. However, this approach can lead to over-coupling, causing a sharp decline in performance when modalities are missing. Even with completed modalities, the performance may fall short compared to simple zero padding, which is unsatisfactory. Moreover, these methods often lack a precise generation process, making it difficult to produce missing modality features that closely align with authentic multimodal feature distribution, 
leading to minimal model improvements.

In this paper, we propose an Attention-based Diffusion model for Missing Modalities feature Completion framework (ADMC), as illustrated in Fig.~\ref{fig_2}. To prevent over-coupling between features and the classification network, we independently train feature extraction networks for each modality through single-modality classification, preserving their unique characteristics. Subsequently, we introduce an Attention-based Diffusion Network (ADN) to generate missing modality features that closely adhere to the authentic multimodal feature distribution. ADN operates in the latent space of full-modality features, training a denoising autoencoder with a self-attention mechanism to capture inter-modal dependencies. In the presence of missing modalities, the module completes them through a sequence of denoising steps. Finally, to ensure optimal performance of available modalities even when some are missing, we train separate multimodal feature fusion networks (MF) for each missing modality scenario. MF employs a self-attention mechanism to calculate the weights for each modality and aggregates them accordingly. Interestingly, we find that even in full-modality scenarios, using ADN for cross-modal generation can provide supplemental information, enhancing the final recognition. Thus, our method excels in both missing and full-modality scenarios, achieving state-of-the-art results on the IEMOCAP and MIntRec benchmarks.

Our contributions can be summarized as follows:
\begin{itemize}

\item We introduce a novel ADMC framework for missing modality completion that integrates independently trained feature extraction networks and an attention-based diffusion model.

\item We propose an Attention-based Diffusion Network (ADN) for capturing inter-modal dependencies and generating accurate missing modality features.

\item Our method excels in both missing and full-modality scenarios, showcasing strong practical applicability.

\item We evaluated our approach on the IEMOCAP and MIntRec datasets, achieving state-of-the-art results.

\end{itemize}

\section{RELATED WORK}

\subsection{Incomplete Multimodal Learning}
To address the issue of missing modalities,  simple strategies like filling missing modality values with zeros \cite{padding} or averages \cite{padding1}. However, these methods lack supervision information, leading to suboptimal performance. Low-rank imputation methods attempt to recover missing data by restoring the low-rank nature of complete multimodal data. For instance, Fan et al. \cite{low-rank1}  minimized the tensor tubal rank to handle various missing modalities. Liang et al. \cite{low-rank} incorporated non-linear functions to capture complex correlations in tensor rank minimization. However, these methods may overlook the complementarity between heterogeneous modalities, resulting in suboptimal results.

Given the powerful feature representation capabilities of deep learning methods, they have shown remarkable performance in reconstructing missing modalities. Early approaches include Autoencoders (AE) \cite{AE}, Generative Adversarial Networks (GAN) \cite{GAN}, and Variational Autoencoders (VAE) \cite{VAE}. Recently, Tran et al. \cite{CRA} proposed a Cascade Residual Autoencoder (CRA) to reconstruct missing modality data. Zuo et al. \cite{IF-MMIN} and Zhao et al. \cite{MMIN} employed CRA and cycle consistency learning to learn joint multimodal representations for addressing missing modality problems, gaining significant attention due to their superior performance. Yuan et al. \cite{Transformer-based} and zeng et al. \cite{TATE} utilized Transformer architectures to integrate cross-modal information and capture inter-modal dependencies, achieving outstanding results in missing modality completion tasks. 
Additionally, Lian et al. \cite{GCNet} introduced Graph Neural Networks (GNN) to capture speaker and temporal dependencies, effectively enabling the reconstruction of missing modalities. 
These methods have significantly advanced research in the field of incomplete multimodal learning.

\subsection{Diffusion Model}

Diffusion models were first applied in the field of image processing. 
Ho et al. \cite{DDPM} significantly improved the quality of generated samples by simplifying the training objective (directly predicting noise) and optimizing model parameterization. Compared to other generative models, such as GANs and VAEs, these improvements demonstrated a more stable training process. Building on this foundation, Song et al. \cite{SBGM} proposed a unified framework for diffusion models by integrating score matching with stochastic differential equations (SDE), addressing the inefficiency of traditional diffusion models. These advancements have established a solid theoretical foundation and practical methods for high-efficiency diffusion models.

With the further development of diffusion models, their conditional generation capabilities have been progressively explored and optimized. Dhariwal et al. \cite{Classifier} and Ho et al. \cite{Classifier-free} leveraged the gradients of pre-trained classifiers or directly trained generative models to achieve conditional control, laying the groundwork for conditional diffusion. 
Conditional diffusion models can also guide the generation process by integrating textual descriptions or other modalities. For example,  Zhong et al. \cite{zhong} combined pre-trained language models with diffusion models to generate high-quality images that align closely with textual descriptions. Similarly, Saharia et al. \cite{palette} applied diffusion models to conditional image generation, demonstrating remarkable performance across various image transformation tasks. 
Furthermore, Liu et al. \cite{audioldm} implemented diffusion processes in latent space for text-to-audio cross-modal generation, significantly reducing computational complexity while preserving generation quality.
These advancements have further expanded the application scenarios of diffusion models.

\begin{figure*}[!t]
\centering
\includegraphics[width=0.85\linewidth]{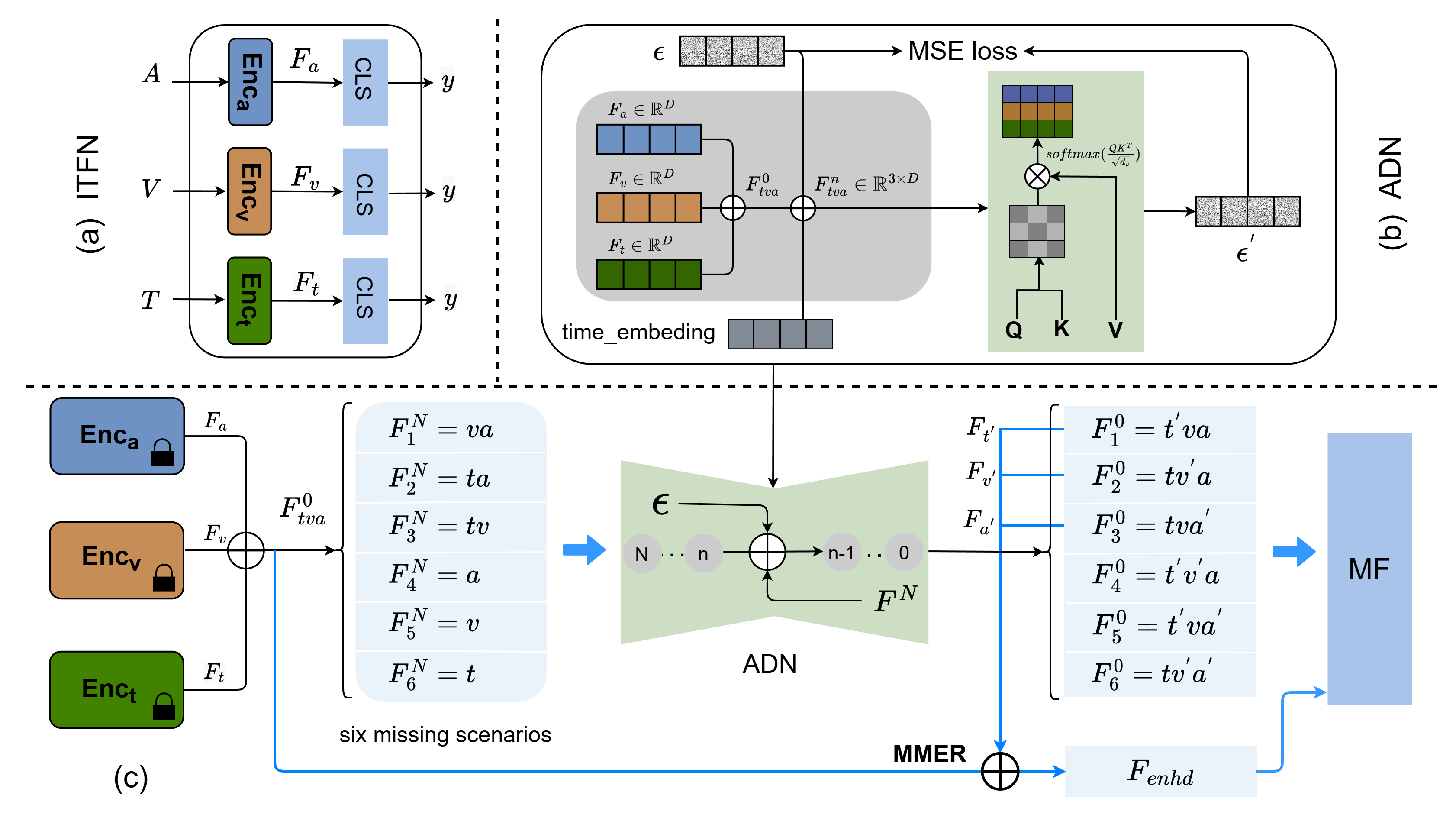}
\caption{Shows the overall framework of the proposed ADMC method. (a) The independently trained feature extraction networks (ITFN) for each modality, aiming to preserve their unique characteristics. (b) The Attention-based Diffusion Network (ADN), aiming to
capture inter-modal dependencies. (c) Multimodal recognition tasks under missing and full-modality scenarios, respectively.
}
\label{fig_2}
\end{figure*}

\section{METHODOLOGY}

Our method aims to classify the category $y$ for each video segment, as shown in Fig.~\ref{fig_2}.
We use \(A\), \(V\), and \(T\) to represent the raw acoustic, visual, and textual datas of the video segment, respectively. 
During the training phase, the model is provided with the full-modalities as input, while in the testing phase, it accepts either partial or full-modalities. 
The method consists of three stages: First, independently train feature extraction networks for each modality to obtain visual  \(F_v\), acoustic \(F_a\), and textual \(F_t\) features (Fig.~\ref{fig_2}(a)).
Next, fix these feature extraction networks (\( Enc_v \) for visual, \( Enc_a \) for acoustic, and \( Enc_t \) for textual) and train an Attention-based Diffusion Network (ADN) to generate missing modalities features (Fig.~\ref{fig_2}(b)). 
Finally, with the parameters of Enc and ADN fixed, we train Multimodal Fusion classification networks (MF) based on the full-modalities, which outputs the final classification results (Fig.~\ref{fig_2}(c)).

\subsection{Modality-specific Feature Extraction}

We follow the feature extraction approach described in \cite{MMIN,IF-MMIN,mintrec}, extracting visual \(F_v\), audio \(F_a\), and text \(F_t\) utterance-level representations from the raw datas.
These features are extracted using networks trained on single-modality classification, with early stopping applied to prevent overfitting, as shown in Fig.~\ref{fig_2}(a). The parameters of these feature extraction networks are fixed during subsequent tasks. For ease of processing, these features are stacked to form a multimodal feature representation \(F^0_{tva}\), which serves as the input for the following modules.

\subsection{Attention-based Diffusion Networks}

The Attention-based Diffusion Networks (ADN) consists of two processes: forward and reverse \cite{audioldm}.

In the forward process, multimodal features \(F^0_{tva}\) are gradually corrupted by adding Gaussian noise \(\epsilon\). As the diffusion step \(n \in [1, \dots, N]\) increases, the original information is progressively lost, approaching a noise distribution, as defined in Eq.~\ref{q_1}. Additionally, a multi-head self-attention network\cite{Attention} is employed to capture inter-modal dependencies and predict the added noise \(\epsilon'\) at each time step, as defined in Eq.~\ref{q_2}. The ADN network is trained by minimizing the mean squared error (MSE) between \(\epsilon\) and \(\epsilon'\), as shown in Fig.~\ref{fig_2}(b).
\begin{equation}
\label{q_1}
   q(F^n_{tva} | F^0_{tva}) = \mathcal{N}(F^n_{tva}; \sqrt{\bar{\alpha}_n} F^0_{tva}, (1 - \bar{\alpha}_n)\epsilon),
\end{equation}
\begin{equation}
\label{q_2}
\epsilon' = \text{Attention}(Q^l, K^l, V^l) \quad \text{for } l = 1, 2, \dots, L,
\end{equation}
Where \( F^n_{tva} \) represents the multimodal feature after the \( n \)-th forward diffusion step, and \(\bar{\alpha}_n\) is the cumulative noise scheduling parameter. \(Q^l\), \(K^l\), and \(V^l\) correspond to the query, key, and value vectors of \( F^n_{tva} \), respectively, where \(L\) indicates the number of layers in the self-attention network.

In the reverse process, we incorporate the available modalities part of the latent representation \(F_{o}\) into the generated latent representation \(F_{tva}\). 
Starting from step \( n \) in the reverse diffusion process, after each inference step (Eq.~\ref{q_3}), we update \( F^{n-1}_{tva} \) using Eq.~\ref{q_4}. 
By using this approach, we can generate the missing modality features while preserving the ground-truth data of the available modalities.
\begin{equation}
\begin{aligned}
\label{q_3}
p_\theta(F^{n-1}_{tva} | F^n_{tva}) &= \mathcal{N}\bigg(F^{n-1}_{tva}; \frac{1}{\sqrt{\alpha_n}} \left( F^n_{tva} - \right. \\
&\ \left. \frac{\beta_n}{\sqrt{1 - \bar{\alpha}_n}} \epsilon_\theta(F^n_{tva}, n) \right), \sigma^2_n \mathbf{I} \bigg),
\end{aligned}
\end{equation}
\begin{equation}
\label{q_4}
F^{n-1}_{tva} = m \odot F^{ n-1}_{o} + (1 - m) \odot F^{n-1}_{tva},
\end{equation}
Where \( F^{n-1}_{tva} \) represents the latent representation after the \( (N-n+1) \)-th reverse diffusion step, while \( F^{n-1}_{o} \) denotes the latent representation of the available modality part after the \( (n-1) \)-th forward diffusion step. 
The optimal \( n \) is determined experimentally. The mask \( m \) indicates missing and available modalities, and \( \beta_n \) is the noise scheduling parameter. 

\subsection{Modality Completion and Enhancement}

We validate our method under both missing modality and full-modality scenarios, as shown in Fig.~\ref{fig_2}(c).
In the missing modality scenario, we replace the missing features with Gaussian noise \(\epsilon\) and use the trained ADN to generate the missing modality features.
This process ensures the completeness of multimodal data, enhancing the model's robustness across various missing modality scenarios.

In the full-modality scenario, we propose a MultiModal Enhancement Recognition (MMER) method. The input consists of full-modality features, we use the ADN for cross-modal generation (e.g., using \( F_a \) and \( F_v \) to generate \( F'_t \)). 
These generated features \( F' \) encode key information from other modalities and can provide additional complementary cues to the original modality.
To integrate this information, we concatenate \( F \) with \( F' \), resulting in the enhanced modality representation \(F_{\text{enhd}}\). 
As shown in Eq.~\ref{q_5}, where \( F \) denotes the original modality feature, and \( F' \) represents the generated cross-modal feature obtained from the ADN. The function \(\text{cat}(\cdot)\) denotes the concatenation operation, performed along the channel dimension.

\begin{equation}
\label{q_5}
F_{\text{enhd}} = \text{cat}(F, F').
\end{equation}

\subsection{Multimodal Feature Fusion Network}

The output feature $F_M \in \mathbb{R}^{M \times D}$ obtained from completion or enhancement operation is processed by a multi-head self-attention based multimodal fusion network (MF). Specifically, MF first employs the multi-head attention mechanism to model inter-modal dependencies and learn attention weights for each modality. Then, it performs a weighted aggregation of modality features followed by mean pooling along the modality dimension to obtain the fused representation $u$, as shown in Eq.~\ref{q_6}. Finally, $u$ is fed into a fully connected classification network to produce the final prediction.  

\begin{equation}
u = \text{Mean}\!\left(\text{softmax}\!\left(\frac{QK^\top}{\sqrt{d_k}}\right)V\right),
\label{q_6}
\end{equation}
where $Q = F_M W^Q$, $K = F_M W^K$, and $V = F_M W^V$. $W^Q$, $W^K$ and $W^V$ are learnable parameters, $d_k$ denotes the dimension of the key vectors, and Mean($\cdot$) represents the mean pooling operation along the modality dimension.

\section{Experiments Setup}
\subsection{Dataset}
We evaluated our proposed method on two benchmark multimodal datasets: IEMOCAP \cite{IEMOCAP} and MIntRec \cite{mintrec}. IEMOCAP is an Interactive Emotional dataset with four emotional labels. MIntRec is a multimodal intent recognition dataset collected from life TV dramas, featuring twenty intent labels.

In the IEMOCAP dataset, the frame-level features are 130-dim OpenSMILE features (IS13 ComParE) \cite{OpenSMILE}, 342-dim Denseface features from a pretrained DenseNet \cite{DenseNet}, and 1024-dim BERT-large embeddings \cite{Bert}. In the MIntRec dataset, the frame-level features are 768-dim wav2vec 2.0 features \cite{wav2vec2}, 256-dim Faster R-CNN features from a pretrained ResNet-50 \cite{mintrec}, and 768-dim BERT embeddings \cite{Bert}.
Similar to \cite{MMIN,IF-MMIN}, for \(Enc_a\), we use an LSTM \cite{LSTM} network to capture temporal information from sequential frame-level acoustic features and apply max-pooling to obtain utterance-level acoustic embeddings. 
For \(Enc_v\), we use the same approach as \(Enc_a\) to obtain utterance-level visual embeddings. 
For \(Enc_t\), we employ a TextCNN \cite{kim2014} to obtain utterance-level textual embeddings from sequential word-level features.

\subsection{Implementation details}

The hidden size for encoders \( Enc_a \) and \( Enc_v \) is set to 256. \( Enc_a \) consists of 3 convolutional blocks with kernel sizes of 3, 4, and 5, each producing an output size of 256. 
In the ADN module, the self-attention network has 4 encoder layers with 8 attention heads and a hidden size of 1024. 
The diffusion model uses noise scheduling parameters from \( \beta_s = 0.0003 \) to \( \beta_e = 0.06 \) over \( N = 1000 \) steps. 
In the MF module, the self-attention network has 2 layers with 4 attention heads, a hidden dimension of 1024, and a dropout rate of 0.5. The classifier consists of 2 fully connected layers of size \{128, 4\}.
We use the Adam optimizer with a dynamic learning rate, starting at 0.0001, and apply the Lambda LR scheduling method for learning rate updates. We select the best model on the validation set and report its performance on the testing set. All models are implemented with Pytorch deep learning toolkit and run  on an RTX 4090 GPU.

\subsection{Baseline}

The method is compared with recent approaches addressing missing modalities: 
{\bf{1) CRA}} \cite{CRA}: is an extension of AE \cite{AE}. It combines a series of residual autoencoders into a cascaded architecture for missing data imputation.
{\bf{2) MCTN}} \cite{MCTN}: Utilizes cyclic translation between modalities to obtain joint representations.
{\bf{3) MMIN}} \cite{MMIN}: Learns joint representations via cross-modality imagination using cascade residual auto-encoder and cycle consistency learning.  
{\bf{4) IF-MMIN}} \cite{IF-MMIN}: Enhances MMIN by introducing modality-invariant features to reduce modality gaps. 
{\bf 5) IMDer}~\cite{IMDer}: A score-based diffusion framework that models the distribution of missing modalities and reconstructs them by denoising random noise under semantic conditions.
{\bf{6) MPLMM}} \cite{MPLMM}: Learning of both intra- and inter-modal information through generative and missing-signal prompts, while significantly reducing the number of trainable parameters.
{\bf{7) P-RMF}} \cite{P-RMF}: Combines quantified modality intrinsic uncertainty to learn stable multimodal joint representations, enhancing the model's robustness to noise and missing data.

\begin{table*}[!h]
\centering
\caption{The experimental results of our method with recent other methods on the IEMOCAP (i.e. testing condition \{T\} indicates that only textual modality is available and both visual and acoustic modalities are missing), Bold numbers indicate the best performance, and methods marked with * have results reproduced using open-source code, while other results are from the original papers.}

\begin{tabular}{lccccccc}
\toprule
\multirow{2}{*}{Methods}& \{A\}& \{V\}& \{T\}& \{AV\}& \{AT\}& \{VT\}&Average\\
\cmidrule{2-8}
& WA$\uparrow$/UA$\uparrow$& WA$\uparrow$/UA$\uparrow$& WA$\uparrow$/UA$\uparrow$& WA$\uparrow$/UA$\uparrow$& WA$\uparrow$/UA$\uparrow$& WA$\uparrow$/UA$\uparrow$ &WA$\uparrow$/UA$\uparrow$\\
\midrule
CRA (2017)& 48.1/49.1& 47.7/48.2& 62.3/63.5& 57.1/54.2& 68.7/70.3& 68.2/67.7&58.7/58.8\\
MCTN (2019) & 49.7/51.6& 48.9/45.7& 62.4/63.7& 56.3/55.8& 68.3/69.4& 67.8/68.3&58.9/59.1\\
MMIN (2021)& 56.5/59.0&52.5/51.6& 66.5/68.0& 63.9/65.4& 72.9/75.1& 72.6/73.6&
64.1/65.2\\

IF-MMIN (2023) & 56.2/58.1& 51.9/50.4& 67.0/68.2& 65.3/66.5& 74.0/75.4& 72.6/73.6&64.5/65.3\\

$\text{IMDer}^{*}$ (2023) &54.1/55.2 &55.6/56.3 & 68.2/68.5 & 67.1/68.3 & 71.2/72.3 & 73.1/73.4 & 64.8/65.7\\

MPLMM (2024) &\textbf{59.7}/- &57.6/- &69.2/- &67.2/- &75.9/- & 74.6/- &67.4/-\\

$\text{P-RMF}^{*}$ (2025) &52.3/56.7 &58.6/55.4& 68.3/69.6 &67.8/68.7 & 68.9/69.9 &73.6/75.4 &64.9/66.0\\

Ours& 59.2/\textbf{61.6}& \textbf{59.8/57.2}& \textbf{71.4/74.1}& \textbf{72.1/72.9}& 74.1/\textbf{76.6}& \textbf{75.2/78.5}&\textbf{68.7/70.2}\\
\toprule
\end{tabular}
\label{label_1}
\end{table*}

\begin{table*}[t]
\centering
\caption{The experimental results of our method with recent other methods on the MIntRec.}
\begin{tabular}{lccccccc}
\toprule
\multirow{2}{*}{Methods}& \{A\}& \{V\}& \{T\}& \{AV\}& \{AT\}& \{VT\}&Average\\
\cmidrule{2-8}
& WA$\uparrow$/UA$\uparrow$& WA$\uparrow$/UA$\uparrow$& WA$\uparrow$/UA$\uparrow$& WA$\uparrow$/UA$\uparrow$&WA$\uparrow$/UA$\uparrow$& WA$\uparrow$/UA$\uparrow$ &WA$\uparrow$/UA$\uparrow$\\
\midrule
CRA (2017)& 28.2/17.8& 11.9/7.1& 68.5/64.6& 26.8/19.4& 68.7/65.3& 67.0/64.1&45.2/39.1\\
MCTN (2019)& 26.5/24.3& 13.7/10.1& 67.9/62.3& 27.7/24.5& 68.2/64.1& 69.0/64.6&45.5/41.7\\
MMIN (2021)& 24.7/14.6& 15.2/6.7& 68.3/\textbf{65.5} & 25.1/15.6  &69.1/65.8& 69.4/65.5 &45.3/39.0\\
IF-MMIN (2023)& 29.2/23.4& 12.8/6.7& 67.6/62.1& 28.1/23.2& 68.3/62.2& 67.2/61.8&45.5/39.9\\

$\text{IMDer}^{*}$ (2023) &23.6/17.3 & 10.6/7.9 &67.3/62.8 & 27.6/19.8 & 69.3/64.2 &68.1/63.6 & 44.4/39.2\\

$\text{P-RMF}^{*}$ (2025) &24.8/16.2  &11.9/5.3 &67.6/64.9 & 23.7/16.9 &70.4/65.3
&67.8/65.3 & 44.6/39.0\\

Ours& \textbf{33.4/29.3}& \textbf{15.4/10.2}& \textbf{71.6/65.7}& \textbf{34.3/29.6}&\textbf{72.3/65.8}& \textbf{71.8}/65.1&\textbf{49.8/44.3}\\
\toprule
\end{tabular}
\label{label_2}
\end{table*}

\section{Experimental results and analysis }

\subsection{Comparative Study}

To comprehensively evaluate the effectiveness of our method, we compare it with previous state-of-the-art approaches and report weighted accuracy (WA) and unweighted accuracy (UA), as shown in Tables~\ref{label_3} and ~\ref{label_4}.
On the IEMOCAP dataset, our results significantly outperform previous approaches across all modality-missing scenarios, with an average performance improvement of 1.9\%. 
Similarly, on the MIntRec dataset, our method sustained these advantages, achieving an average performance improvement of 9.4\%.
These results demonstrate the robustness and generalization of our method in missing modality completion tasks.

Specifically, compared to autoencoder-based methods (AE, CRA), translation-based methods (MCTN, MMIN) typically yield better performance. The modality-to-modality transformations in these methods help capture inter-modality dependencies and generate features for missing modalities. However, IF-MIN and P-RMF consistently underperform ADMC in single-modality input scenarios. This performance gap arises primarily from their joint training strategies, which lead to strong coupling between the feature extraction networks of different modalities. When certain modalities are missing, this coupling causes a significant drop in model performance.
Although IMDer also uses diffusion models to reconstruct missing modalities, its performance is limited by the lack of an attention-based cross-modal mechanism.
In contrast, ADMC utilizes independently trained feature extraction networks (ITFN), which effectively reduce inter-modality coupling. This approach allows each available modality to fully express its representational power, even when some modalities are missing. Additionally, ADMC employs Attention-based Diffusion Networks (ADN) to capture inter-modality dependencies and generate missing modality features that closely align with the authentic multimodal feature distribution, thereby significantly enhancing the overall performance of the model.

\subsection{Ablation Study}

To validate the contributions of the independently trained feature extraction networks (ITFN) and ADN modules, we conducted ablation experiments, as shown in Tables~\ref{label_3} and ~\ref{label_4}. 

\begin{table}[t]
\centering
\caption{Ablation study results on IEMOCAP.}
\begin{tabular}{c@{\hskip 6pt}c@{\hskip 6pt}c@{\hskip 6pt}c@{\hskip 6pt}c@{\hskip 6pt}c@{\hskip 6pt}c@{\hskip 6pt}c}
\toprule
\multirow{2}{*}{Methods}& \{A\}& \{V\}& \{T\}& \{AV\}& \{AT\}& \{VT\}&Average\\
\cmidrule{2-8}
& WA$\uparrow$& WA$\uparrow$& WA$\uparrow$& WA$\uparrow$& WA$\uparrow$& WA$\uparrow$ &WA$\uparrow$\\
\midrule

w/o ITFN& 57.0& 54.7& 69.2& 67.8& \textbf{75.6}& 74.5&66.5\\

w/o ADN& 58.0& 58.1& 70.0& 70.2& 73.9& 74.3&67.6\\

W/ U-Net& 58.2& 57.6& 71.4& 71.3& 72.7& 74.5&67.5\\

Ours& \textbf{59.2}& \textbf{59.8}& \textbf{71.4}& \textbf{72.1}& 74.1& \textbf{75.2}&\textbf{68.7}\\
\toprule
\end{tabular}
\label{label_3}
\end{table}

\begin{table}[t]
\centering
\caption{Ablation study results on the MIntRec.}
\begin{tabular}{c@{\hskip 6pt}c@{\hskip 6pt}c@{\hskip 6pt}c@{\hskip 6pt}c@{\hskip 6pt}c@{\hskip 6pt}c@{\hskip 6pt}c}
\toprule
\multirow{2}{*}{Methods}& \{A\}& \{V\}& \{T\}& \{AV\}& \{AT\}& \{VT\}&Average\\
\cmidrule{2-8}
& WA$\uparrow$& WA$\uparrow$& WA$\uparrow$& WA$\uparrow$
&WA$\uparrow$& WA$\uparrow$ &WA$\uparrow$\\
\midrule
w/o ITFN& 31.6& 11.7& 71.1& 31.8& 71.1& 71.3 &48.1\\

w/o ADN& 32.4& 15.0& 71.5& 33.7&71.5& 71.3 &49.3\\

w/ U-Net& 32.2& 14.5& 71.5& 33.6& 71.1& 71.5&49.0\\

Ours& \textbf{33.4}& \textbf{15.4}& \textbf{71.6}& \textbf{34.3}&\textbf{72.3}& \textbf{71.8}&\textbf{49.8}\\
\toprule
\end{tabular}
\label{label_4}
\end{table}

{\bf 1) w/o ITFN:} Removing the independently trained feature extraction networks and using the jointly trained extractor, while missing modality features are generated by the ADN, leads to a performance drop under missing-modality conditions (e.g., using only audio or only visual modality). This indicates that joint training leads to over-coupled feature extraction across modalities, making the model less robust when certain modalities are missing.
{\bf 2) w/o ADN:} Retaining the independent feature extractors but replacing missing modality features with zero vectors instead of reconstructing them via ADN causes performance degradation across all scenarios, showing that without reconstructing missing modalities, multimodal information cannot be fully utilized. With ADN, the model can generate complementary features based on the available modalities, thereby improving robustness under missing-modality conditions.
{\bf 3) w/ U-Net:} Replacing the self-attention module in ADN with a standard U-Net\cite{u-net} yields lower performance. Although U-Net can generate features close to the target distribution, it lacks explicit modeling of cross-modal dependencies, resulting in reconstructed features that are less consistent with the global multimodal semantics and thus lower overall performance. In contrast, ADN captures cross-modal relationships through self-attention and produces more discriminative complementary features.

\subsection{Visualization Analysis} 

\begin{figure}[!ht]
\centering
\begin{subfigure}[t]{1\linewidth}
    \centering
    \includegraphics[width=1.0\textwidth]{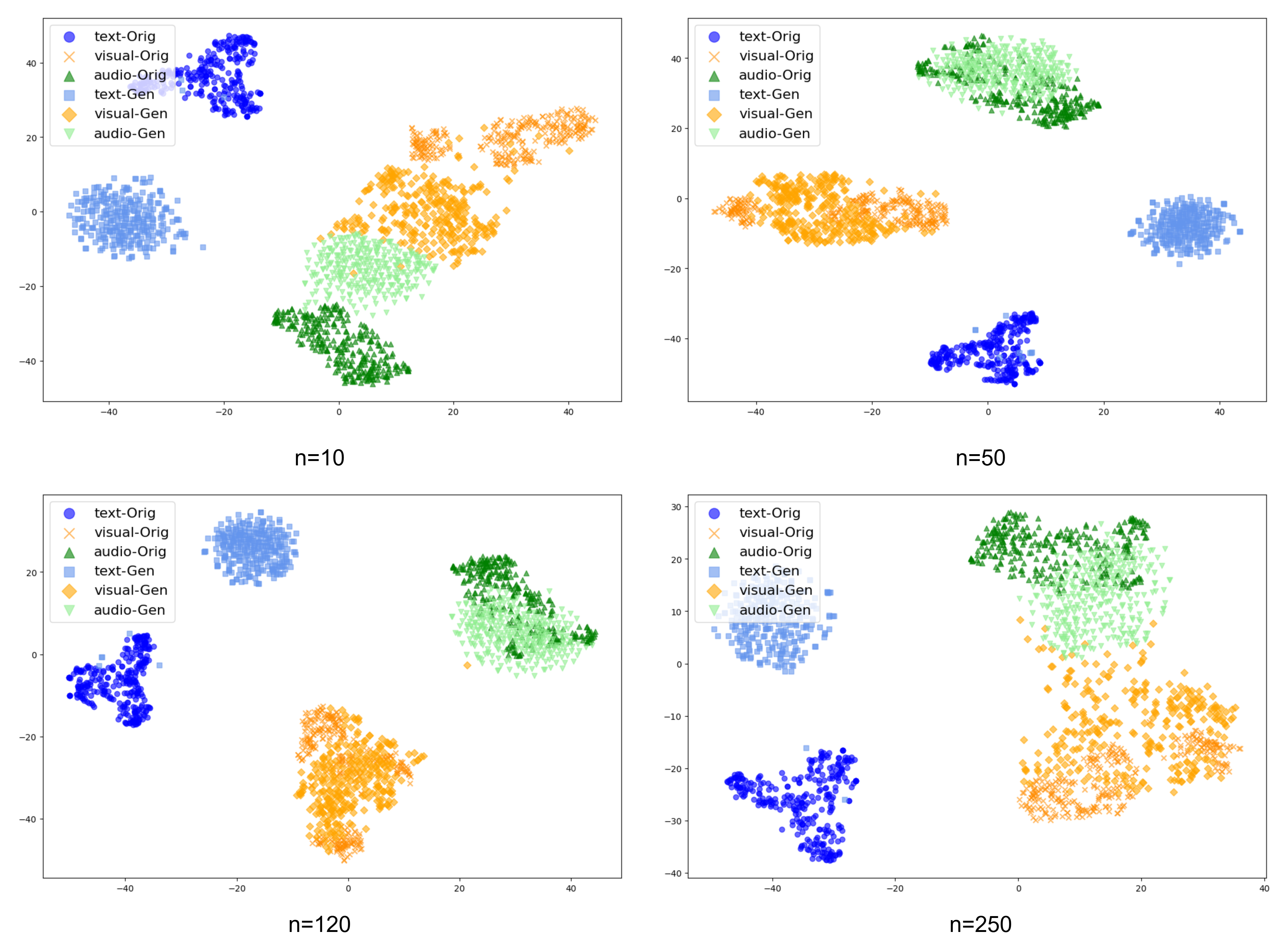}
    \caption{Visualization on the IEMOCAP dataset.}  
    \label{fig_first_case}
\end{subfigure}%


\begin{subfigure}[t]{1\linewidth}
    \centering
    \includegraphics[width=1.0\linewidth]{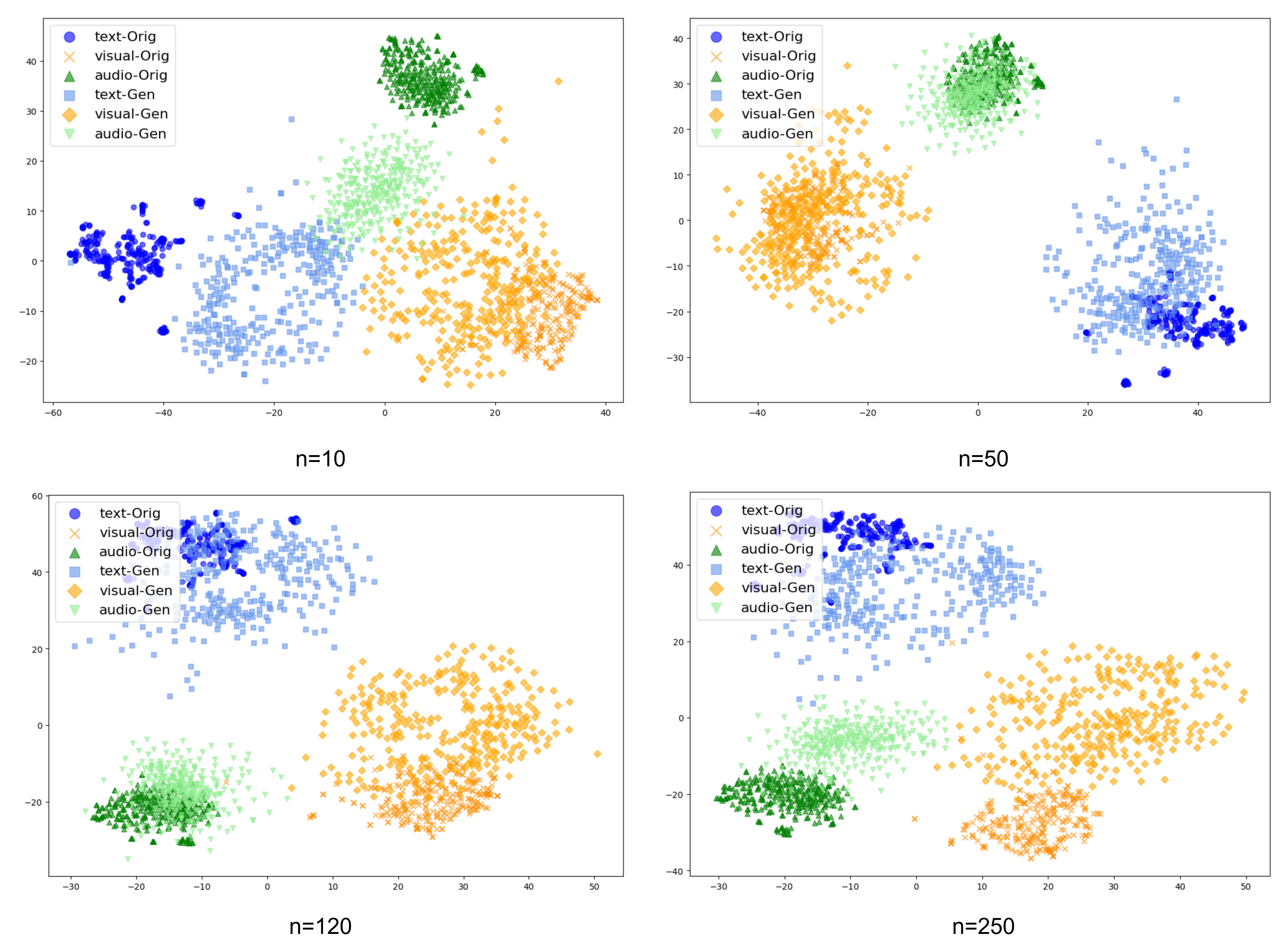}
    \caption{Visualization on the MIntRec dataset.}  
    \label{fig_second_case}
\end{subfigure}
\caption{
Shows the visualization of ground-truth multimodal embeddings and generated embeddings on the IEMOCAP and MIntRec dataset. Blue represents text feature embeddings, green represents audio feature embeddings, and yellow represents visual feature embeddings.
}
\label{fig_3}
\end{figure}

To intuitively analyze the impact of different values of \(n\) on the quality of generated features during reverse generation, we visualized the differences between ground-truth multimodal embeddings and the corresponding generated embeddings. Specifically, we randomly selected 600 real samples and their generated counterparts from each modality, and applied the T-SNE toolkit \cite{TSNE} for dimensionality reduction. As shown in Figure~\ref{fig_3}, we observe that:  
1) Features from different modalities exhibit clear clustering patterns in the embedding space, indicating that the ground-truth embeddings effectively preserve modality-specific information.  
2) Extremely small or large values of \(n\) negatively impact the quality of the generated embeddings, leading to blurred distribution boundaries and overlaps between modalities.  
3) When \(n\) = 50, the generated embeddings closely align with the distribution of the ground-truth embeddings, demonstrating the effectiveness of our method in generating missing modality features. 
4) On the MIntRec dataset, the generated features show a higher degree of dispersion, primarily due to the limited sample size, which restricts the model’s ability to capture the underlying data distribution.

\subsection{Effects of MMER}

To evaluate the effectiveness of the MMER module, we conducted comparative experiments under full-modality scenarios, as shown in Table~\ref{label_5}. 
The experiment compared three methods: unrestricted feature extraction networks (Jointly\_training), fixed feature extraction networks (Use\_ITFN), and using generated features to enhance original modality representations (MMER). 
The results indicate that the Jointly\_Training method outperforms the Use\_ITFN method by dynamically adjusting the feature representations of each modality, enabling the model to achieve optimal performance.
The MMER method achieves the best performance by enhancing original modality representations using generated features.

Existing methods for addressing missing modality problems often perform poorly in full-modality scenarios. In contrast, our method not only focuses on tackling missing modality issues but also demonstrates excellent performance in full-modality tasks. 

\begin{table}[h!]
\centering
\caption{Result of the study on the MMER.}
\begin{tabular}{ccccc}
\toprule
\multirow{2}{*}{Methods}& \multicolumn{2}{c}{IEMOCAP} & \multicolumn{2}{c}{MIntRec} \\
\cmidrule{2-5}
 & WA$\uparrow$& UA$\uparrow$& WA$\uparrow$& UA$\uparrow$\\
\midrule
Jointly\_training&\textbf{77.4}& 79.1 & 71.2 & 65.0
\\

Use\_ITFN& 76.5& 78.1& 71.5 & 65.0
\\

MMER& 77.3& \textbf{79.5} & \textbf{72.5} & \textbf{65.4}\\
\toprule
\end{tabular}
\label{label_5}
\end{table}

\subsection{Experimental of Different Missing Rate}

\begin{figure}[!h]
    \centering
    \includegraphics[width=1\linewidth]{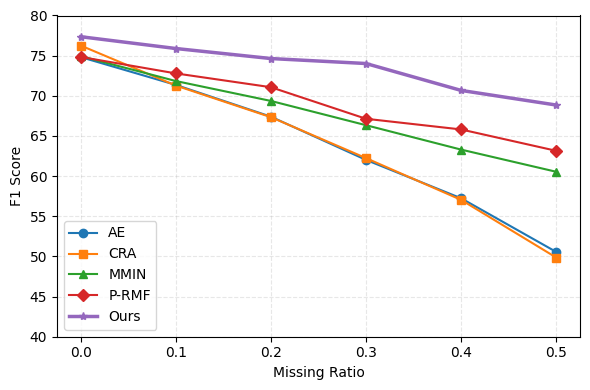}
	\caption{Results under different missing rates on the IEMOCAP dataset.}
	\label{fig_4}
\end{figure}

To evaluate the robustness of multimodal models under inter-modality missingness, we using a probabilistic approach to randomly drop one or two modalities at varying missing ratios \(  p \in {0.1, 0.2, \ldots, 0.5} \). 
The performance trends of different models under increasing missing ratios are shown in Fig.~\ref{fig_4}.

As the missing ratio increases, the F1 scores of all baseline methods consistently decline, indicating that inter-modality missingness disrupts semantic integrity and weakens the effectiveness of multimodal feature fusion. 
In contrast, our method achieves the best performance across all missing ratios and suffers the smallest performance degradation. This advantage stems, on the one hand, from the ADN module, which generates missing modality features that closely match the real feature distribution and thus enhance multimodal fusion, and on the other hand from the decoupled feature extraction mechanism in our framework, which effectively mitigates the impact of missing modalities on overall model performance.

\subsection{Time Complexity and Inference Cost}

Introducing a diffusion model inevitably increases inference time. We therefore analyze the computational complexity and practical inference cost of the proposed ADN module. ADN is a 4-layer Transformer-encoder-based self-attention network. For an input with \(L\) modalities and feature dimension \(D\), the per-step inference complexity is \(\mathcal{O}(L^2 D + L D^2)\). In our setting, the number of modalities is very small (\(L = 3\)), so the overall complexity is substantially lower than that of standard long-sequence Transformers.

We further evaluate the actual inference latency on an RTX 4090 GPU. With 1{,}000 diffusion denoising steps, the additional inference time introduced by ADN is approximately 50 ms per sample, which is acceptable for our target scenarios. Moreover, existing acceleration techniques for diffusion models (e.g., DDIM, reduced-step samplers, and step-skipping strategies) can reduce the number of steps to 50–100 with negligible performance degradation, indicating clear potential for further efficiency improvements.

\section{CONCLUSION}

In this paper, we propose an Attention-based Diffusion Model for Missing modalities feature Completion (ADMC), which integrates independently trained feature extraction networks and an Attention-based Diffusion Network (ADN). 
Our approach preserves the distinct characteristics of each modality while generating missing modality features that closely adhere to the authentic multimodal feature distribution. 
Extensive experimental results demonstrate that our method achieves state-of-the-art performance.
In future work, we will continue to explore the applications of attention-based diffusion networks in a broader range of tasks.

\section{Safe and Responsible Innovation Statement}

This work supports more robust multimodal emotion and intent recognition under missing-modality conditions, which may improve accessibility and reliability in human-computer interaction. However, such systems may also introduce risks related to privacy, bias, and misuse. The datasets used in this study are public benchmarks, and we do not collect new personal data. We acknowledge that emotion and intent recognition models can inherit demographic, cultural, and contextual biases from training data, potentially leading to unfair outcomes. Our method is intended as a decision-support tool rather than a sole basis for high-stakes decisions. Responsible deployment should include human oversight, bias evaluation across groups, privacy protection, and clear usage boundaries.


\bibliographystyle{ACM-Reference-Format}
\bibliography{sample-base}

@inproceedings{EMMR,
  title={Mitigating inconsistencies in multimodal sentiment analysis under uncertain missing modalities},
  author={Zeng, Jiandian and Zhou, Jiantao and Liu, Tianyi},
  booktitle={Proceedings of the 2022 Conference on Empirical Methods in Natural Language Processing},
  pages={2924--2934},
  year={2022}
}

@inproceedings{adversarial,
  title={Deep adversarial learning for multi-modality missing data completion},
  author={Cai, Lei and Wang, Zhengyang and Gao, Hongyang and Shen, Dinggang and Ji, Shuiwang},
  booktitle={Proceedings of the 24th ACM SIGKDD international conference on knowledge discovery \& data mining},
  pages={1158--1166},
  year={2018}
}

@inproceedings{mintrec,
  title={Mintrec: A new dataset for multimodal intent recognition},
  author={Zhang, Hanlei and Xu, Hua and Wang, Xin and Zhou, Qianrui and Zhao, Shaojie and Teng, Jiayan},
  booktitle={Proceedings of the 30th ACM International Conference on Multimedia},
  pages={1688--1697},
  year={2022}
}

@article{ref5,
  title={An effective multimodal representation and fusion method for multimodal intent recognition},
  author={Huang, Xuejian and Ma, Tinghuai and Jia, Li and Zhang, Yuanjian and Rong, Huan and Alnabhan, Najla},
  journal={Neurocomputing},
  volume={548},
  pages={126373},
  year={2023},
  publisher={Elsevier}
}

@inproceedings{ref6,
  title={Token-level contrastive learning with modality-aware prompting for multimodal intent recognition},
  author={Zhou, Qianrui and Xu, Hua and Li, Hao and Zhang, Hanlei and Zhang, Xiaohan and Wang, Yifan and Gao, Kai},
  booktitle={Proceedings of the AAAI Conference on Artificial Intelligence},
  volume={38},
  number={15},
  pages={17114--17122},
  year={2024}
}

@inproceedings{MCTN,
  title={Found in translation: Learning robust joint representations by cyclic translations between modalities},
  author={Pham, Hai and Liang, Paul Pu and Manzini, Thomas and Morency, Louis-Philippe and P{\'o}czos, Barnab{\'a}s},
  booktitle={Proceedings of the AAAI conference on artificial intelligence},
  volume={33},
  number={01},
  pages={6892--6899},
  year={2019}
}

@inproceedings{TransM,
  title={Transmodality: An end2end fusion method with transformer for multimodal sentiment analysis},
  author={Wang, Zilong and Wan, Zhaohong and Wan, Xiaojun},
  booktitle={Proceedings of the web conference 2020},
  pages={2514--2520},
  year={2020}
}

@inproceedings{Transformer-based,
  title={Transformer-based feature reconstruction network for robust multimodal sentiment analysis},
  author={Yuan, Ziqi and Li, Wei and Xu, Hua and Yu, Wenmeng},
  booktitle={Proceedings of the 29th ACM International Conference on Multimedia},
  pages={4400--4407},
  year={2021}
}

@inproceedings{TATE,
  title={Tag-assisted multimodal sentiment analysis under uncertain missing modalities},
  author={Zeng, Jiandian and Liu, Tianyi and Zhou, Jiantao},
  booktitle={Proceedings of the 45th International ACM SIGIR Conference on Research and Development in Information Retrieval},
  pages={1545--1554},
  year={2022}
}

@inproceedings{MMIN,
  title={Missing modality imagination network for emotion recognition with uncertain missing modalities},
  author={Zhao, Jinming and Li, Ruichen and Jin, Qin},
  booktitle={Proceedings of the 59th Annual Meeting of the Association for Computational Linguistics and the 11th International Joint Conference on Natural Language Processing (Volume 1: Long Papers)},
  pages={2608--2618},
  year={2021}
}

@inproceedings{IF-MMIN,
  title={Exploiting modality-invariant feature for robust multimodal emotion recognition with missing modalities},
  author={Zuo, Haolin and Liu, Rui and Zhao, Jinming and Gao, Guanglai and Li, Haizhou},
  booktitle={ICASSP 2023-2023 IEEE International Conference on Acoustics, Speech and Signal Processing (ICASSP)},
  pages={1--5},
  year={2023},
  organization={IEEE}
}

@article{IEMOCAP,
  title={IEMOCAP: Interactive emotional dyadic motion capture database},
  author={Busso, Carlos and Bulut, Murtaza and Lee, Chi-Chun and Kazemzadeh, Abe and Mower, Emily and Kim, Samuel and Chang, Jeannette N and Lee, Sungbok and Narayanan, Shrikanth S},
  journal={Language resources and evaluation},
  volume={42},
  pages={335--359},
  year={2008},
  publisher={Springer}
}

@inproceedings{OpenSMILE,
  title={Opensmile: the munich versatile and fast open-source audio feature extractor},
  author={Eyben, Florian and W{\"o}llmer, Martin and Schuller, Bj{\"o}rn},
  booktitle={Proceedings of the 18th ACM international conference on Multimedia},
  pages={1459--1462},
  year={2010}
}

@inproceedings{DenseNet,
  title={Densely connected convolutional networks},
  author={Huang, Gao and Liu, Zhuang and Van Der Maaten, Laurens and Weinberger, Kilian Q},
  booktitle={Proceedings of the IEEE conference on computer vision and pattern recognition},
  pages={4700--4708},
  year={2017}
}

@article{Bert,
  title={Bert: Pre-training of deep bidirectional transformers for language understanding. arXiv},
  author={Devlin, Jacob and Chang, Ming-Wei and Lee, Kenton and Toutanova, Kristina},
  journal={arXiv preprint arXiv:1810.04805},
  year={2019},
  publisher={Retrieved 2023-01-17, from http://arxiv. org/abs/1810.04805}
}

@article{wav2vec2,
  title={wav2vec 2.0: A framework for self-supervised learning of speech representations},
  author={Baevski, Alexei and Zhou, Yuhao and Mohamed, Abdelrahman and Auli, Michael},
  journal={Advances in neural information processing systems},
  volume={33},
  pages={12449--12460},
  year={2020}
}

@article{LSTM,
  title={Long short-term memory based recurrent neural network architectures for large vocabulary speech recognition},
  author={Sak, Ha{\c{s}}im and Senior, Andrew and Beaufays, Fran{\c{c}}oise},
  journal={arXiv preprint arXiv:1402.1128},
  year={2014}
}

@inproceedings{kim2014,
  title={Convolutional neural networks for sentence classification},
  author={Kim, Yoon},
  booktitle={Proceedings of the 2014 Conference on Empirical Methods in Natural Language Processing (EMNLP)},
  pages={1746--1751},
  year={2014},
  organization={Association for Computational Linguistics}
}

@inproceedings{CRA,
  title={Missing Modalities Imputation via Cascaded Residual Autoencoder},
  author={ Tran, Luan  and  Liu, Xiaoming  and  Zhou, Jiayu  and  Jin, Rong },
  booktitle={Computer Vision \& Pattern Recognition},
  pages={4971-4980},
  year={2017},
}

@article{audioldm,
  title={AudioLDM: Text-to-Audio Generation with Latent Diffusion Models},
  author={Liu, Haohe and Chen, Zehua and Yuan, Yi and Mei, Xinhao and Liu, Xubo and Mandic, Danilo and Wang, Wenwu and Plumbley, Mark D},
  journal={arXiv preprint arXiv:2301.12503},
  year={2023}
}

@article{EmoCaps,
  title={EmoCaps: Emotion capsule based model for conversational emotion recognition},
  author={Li, Zaijing and Tang, Fengxiao and Zhao, Ming and Zhu, Yusen},
  journal={arXiv preprint arXiv:2203.13504},
  year={2022}
}

@article{TSNE,
  title={Visualizing Data using t-SNE},
  author={Van der Maaten, Laurens and Hinton, Geoffrey},
  journal={Journal of Machine Learning Research},
  volume={9},
  pages={2579--2605},
  year={2008}
}

@inproceedings{AE,
  author    = {Duan, Y. and Lv, Y. and Kang, W. and Zhao, Y.},
  title     = {A deep learning based approach for traffic data imputation},
  booktitle = {Proceedings of the 17th International IEEE Conference on Intelligent Transportation Systems (ITSC)},
  pages     = {912--917},
  year      = {2014},
  publisher = {IEEE}
}

@article{IMDer,
  author    = {Wang, Yuanzhi and Li, Yong and Cui, Zhen},
  title     = {Incomplete Multimodality-Diffused Emotion Recognition},
  journal   = {Advances in Neural Information Processing Systems},
  volume    = {36},
  pages     = {17117--17128},
  year      = {2023}
}

@inproceedings{Attention,
  author    = {Vaswani, Ashish and Shazeer, Noam and Parmar, Niki and Uszkoreit, Jakob and Jones, Llion and Gomez, Aidan N. and Kaiser, {\L}ukasz and Polosukhin, Illia},
  title     = {Attention is All You Need},
  booktitle = {Advances in Neural Information Processing Systems},
  volume    = {30},
  year      = {2017}
}

@inproceedings{u-net,
  title={U-net: Convolutional networks for biomedical image segmentation},
  author={Ronneberger, Olaf and Fischer, Philipp and Brox, Thomas},
  booktitle={International Conference on Medical image computing and computer-assisted intervention},
  pages={234--241},
  year={2015},
  organization={Springer}
}

@inproceedings{P-RMF,
  title={Proxy-Driven Robust Multimodal Sentiment Analysis with Incomplete Data},
  author={Zhu, Aoqiang and Hu, Min and Wang, Xiaohua and Yang, Jiaoyun and Tang, Yiming and An, Ning},
  booktitle={Proceedings of the 63rd Annual Meeting of the Association for Computational Linguistics (Volume 1: Long Papers)},
  pages={22123--22138},
  year={2025}
}

@article{MPLMM,
  author    = {Guo, Z. and Jin, T. and Zhao, Z.},
  title     = {Multimodal prompt learning with missing modalities for sentiment analysis and emotion recognition},
  journal   = {arXiv preprint arXiv:2407.05374},
  year      = {2024}
}

@inproceedings{padding,
  title={Training strategies to handle missing modalities for audio-visual expression recognition},
  author={Parthasarathy, Srinivas and Sundaram, Shiva},
  booktitle={Companion Publication of the 2020 International Conference on Multimodal Interaction},
  pages={400--404},
  year={2020}
}

@article{padding1,
  title={Deep partial multi-view learning},
  author={Zhang, Changqing and Cui, Yajie and Han, Zongbo and Zhou, Joey Tianyi and Fu, Huazhu and Hu, Qinghua},
  journal={IEEE transactions on pattern analysis and machine intelligence},
  volume={44},
  number={5},
  pages={2402--2415},
  year={2020},
  publisher={IEEE}
}

@article{low-rank1,
  title={Hyperspectral image restoration using low-rank tensor recovery},
  author={Fan, Haiyan and Chen, Yunjin and Guo, Yulan and Zhang, Hongyan and Kuang, Gangyao},
  journal={IEEE Journal of Selected Topics in Applied Earth Observations and Remote Sensing},
  volume={10},
  number={10},
  pages={4589--4604},
  year={2017},
  publisher={IEEE}
}

@article{low-rank,
  title={Learning representations from imperfect time series data via tensor rank regularization. arXiv 2019},
  author={Liang, PP and Liu, Z and Tsai, YHH and Zhao, Q and Salakhutdinov, R and Morency, LP},
  year = {2019},
  journal={arXiv preprint arXiv:1907.01011}
}

@inproceedings{GAN,
  title={Deep adversarial learning for multi-modality missing data completion},
  author={Cai, Lei and Wang, Zhengyang and Gao, Hongyang and Shen, Dinggang and Ji, Shuiwang},
  booktitle={Proceedings of the 24th ACM SIGKDD international conference on knowledge discovery \& data mining},
  pages={1158--1166},
  year={2018}
}

@inproceedings{VAE,
  title={Semi-supervised deep generative modelling of incomplete multi-modality emotional data},
  author={Du, Changde and Du, Changying and Wang, Hao and Li, Jinpeng and Zheng, Wei-Long and Lu, Bao-Liang and He, Huiguang},
  booktitle={Proceedings of the 26th ACM international conference on Multimedia},
  pages={108--116},
  year={2018}
}

@article{GCNet,
  title={GCNet: Graph completion network for incomplete multimodal learning in conversation},
  author={Lian, Zheng and Chen, Lan and Sun, Licai and Liu, Bin and Tao, Jianhua},
  journal={IEEE Transactions on pattern analysis and machine intelligence},
  volume={45},
  number={7},
  pages={8419--8432},
  year={2023},
  publisher={IEEE}
}

@inproceedings{palette,
  title={Palette: Image-to-image diffusion models},
  author={Saharia, Chitwan and Chan, William and Chang, Huiwen and Lee, Chris and Ho, Jonathan and Salimans, Tim and Fleet, David and Norouzi, Mohammad},
  booktitle={ACM SIGGRAPH 2022 conference proceedings},
  pages={1--10},
  year={2022}
}

@inproceedings{zhong,
  title={Sur-adapter: Enhancing text-to-image pre-trained diffusion models with large language models},
  author={Zhong, Shanshan and Huang, Zhongzhan and Wen, Weushao and Qin, Jinghui and Lin, Liang},
  booktitle={Proceedings of the 31st ACM International Conference on Multimedia},
  pages={567--578},
  year={2023}
}

@article{Classifier-free,
  title={Classifier-free diffusion guidance},
  author={Ho, Jonathan and Salimans, Tim},
  journal={arXiv preprint arXiv:2207.12598},
  year={2022}
}

@inproceedings{Classifier,
  title={Improved denoising diffusion probabilistic models},
  author={Nichol, Alexander Quinn and Dhariwal, Prafulla},
  booktitle={International conference on machine learning},
  pages={8162--8171},
  year={2021},
  organization={PMLR}
}

@article{DDPM,
  title={Denoising diffusion probabilistic models},
  author={Ho, Jonathan and Jain, Ajay and Abbeel, Pieter},
  journal={Advances in neural information processing systems},
  volume={33},
  pages={6840--6851},
  year={2020}
}

@misc{SBGM,
      title={Score-Based Generative Modeling through Stochastic Differential Equations}, 
      author={Yang Song and Jascha Sohl-Dickstein and Diederik P. Kingma and Abhishek Kumar and Stefano Ermon and Ben Poole},
      year={2021},
      eprint={2011.13456},
      archivePrefix={arXiv},
      primaryClass={cs.LG},
      url={https://arxiv.org/abs/2011.13456}, 
}










\end{document}